\documentclass[runningheads]{llncs}  

\usepackage{graphicx}
\usepackage{multirow}
\usepackage{amsfonts,amssymb,amsmath}
\usepackage{float}
\usepackage{comment}
\usepackage{pdfpages}

\usepackage{bbding}

\usepackage[pagebackref=False,breaklinks=true,colorlinks=true,bookmarks=false,citecolor=blue,urlcolor=black,linkcolor=red]{hyperref}

\title{
Future Frame Prediction for Robot-assisted Surgery
}
\titlerunning{Future Frame Prediction in Robotic Surgery}
\institute{}
\author{Xiaojie Gao\inst{1}\orcidID{0000-0001-6517-7630}\and
	Yueming Jin  \inst{1}\orcidID{0000-0003-3775-3877} \and \\
	Zixu Zhao\inst{1}\orcidID{0000-0001-9399-3475} \and
	Qi Dou\inst{1,2}\Envelope\orcidID{0000-0002-3416-9950} \and \\
	Pheng-Ann Heng\inst{1,2}\orcidID{0000-0003-3055-5034}}

\authorrunning{X. Gao et al.}

\institute{Department of Computer Science and Engineering,\\
	The Chinese University of Hong Kong, Hong Kong, China \\ \and
	T Stone Robotics Institute, CUHK, Hong Kong, China \\
	\email{\{xjgao, ymjin, zxzhao, qdou, pheng\}@cse.cuhk.edu.hk} 
}

\begin{document}

\maketitle

\begin{abstract}

Predicting future frames for robotic surgical video is an interesting, important yet extremely challenging problem, given that the operative tasks may have complex dynamics.
Existing approaches on future prediction of natural videos were based on either deterministic models or stochastic models, including deep recurrent neural networks, optical flow, and latent space modeling. 
However, the potential in predicting meaningful movements of robots with dual arms in surgical scenarios has not been tapped so far, which is typically more challenging than forecasting independent motions of one arm robots in natural scenarios. 
In this paper, we propose a ternary prior guided variational autoencoder (TPG-VAE) model for future frame prediction in robotic surgical video sequences. 
Besides content distribution, our model learns motion distribution, which is novel to handle the small movements of surgical tools.
Furthermore, we add the invariant prior information from the gesture class into the generation process to constrain the latent space of our model. 
To our best knowledge, this is the first time that the future frames of dual arm robots are predicted considering their unique characteristics relative to general robotic videos. 
Experiments demonstrate that our model gains more stable and realistic future frame prediction scenes with the suturing task on the public JIGSAWS dataset.


\end{abstract}

\section{Introduction}

With advancements in robot-assisted surgeries, visual data are crucial for surgical context awareness that promotes operation reliability and patient safety. 
Owing to their rich representations and direct perceptibility, surgical videos have been playing an essential role on driving automation process on tasks with obvious clinical significance such as surgical process monitoring~\cite{bhatia2007real,bricon2007context}, gesture and workflow recognition~\cite{funke2019using,gao2020automatic,jin2020multi}, and surgical instrument detection and segmentation~\cite{milletari2018cfcm,colleoni2019deep,islam2019real,jin2019incorporating}. 
However, these methods either only focus on providing semantic descriptions of current situations or directly assume that future information is available, thus developing future scene prediction models for surgeries should be investigated. 
With only happened events available, future prediction will serve as a crucial prerequisite procedure to facilitate advanced tasks including generating alerts~\cite{bhatia2007real}, boosting online recognition with additional information~\cite{twinanda2016endonet}, and supporting decision making of reinforcement learning agents~\cite{liu2018deep,gao2020automatic}. 
Besides supplying an extra source of references, predicted frames could also be converted into an entity demonstration in an imitation style~\cite{tanwani2020motion2vec}, which will indeed accelerate the training of surgeons.

Recently, deep learning techniques have been applied to solve nature video prediction problems. 
To model the temporal dependencies, early methods used the Long Short Term Memory (LSTM)~\cite{hochreiter1997long} or convolutional LSTM, \cite{xingjian2015convolutional} to capture the temporal dynamics in videos~\cite{srivastava2015unsupervised,villegas2017decomposing}. 
Villegas et al. utilized the predicted high-level structures in videos to help generate long-term future frames~\cite{villegas2017learning}.
Methods based on explicit decomposition or multi-frequency analysis were also explored~\cite{villegas2017decomposing,denton2017unsupervised,jin2020exploring}. 
Action-conditional video prediction methods were developed for reinforcement learning agents~\cite{oh2015action,finn2016unsupervised}.
However, these deterministic methods might yield blurry generations due to the missing considerations of multiple moving tendencies~\cite{denton2018stochastic}.
Thus, stochastic video prediction models are proposed to capture the full distributions of uncertain future frames, which are mainly divided into three types: autoregressive models~\cite{kalchbrenner2017video}, flow-based generative models~\cite{kumar2020videoflow}, generative adversarial networks~\cite{tulyakov2018mocogan}, and VAE based approaches~\cite{babaeizadeh2018stochastic,denton2018stochastic}.
As a famous VAE method, Denton and Fergus proposed SVG-LP that uses learned prior from content information rather than standard Gaussian prior~\cite{denton2018stochastic}. 
Diverse future frames could also be generated based on stochastic high-level keypoints~\cite{minderer2019unsupervised,kim2019unsupervised}. 
Although these approaches gain favorable results on general robotic videos~\cite{finn2016unsupervised,babaeizadeh2018stochastic,denton2018stochastic,kumar2020videoflow}, the domain knowledge in dual arm robotic surgical videos, such as the limited inter-class variance and class-dependent movements, are not considered.

Contrast to general robotic videos, a complete surgical video consists of several sub-phases with limited inter-class variances, meaningful movements of specific sub-tasks, and more than one moving instruments, which make this task extremely challenging.
Another challenge in surgical robotic videos is that some actions have more complicated motion trajectories rather than repeated or random patterns in general videos. 
Although predicting one-arm robotic videos were investigated in~\cite{finn2016unsupervised,babaeizadeh2018stochastic}, robots with more than one arm will lead to more diverse frames and make the task even more difficult.
Targeting at the intricate movements of robotic arms, we utilize the content and motion information jointly to assist in predicting the future frames of dual arm robots. 
In addition, experienced surgeons will refer to the classes of current gestures as prior knowledge constantly to forecast future operations. 
How to incorporate content, motion, and class information into the generation model is of great importance to boosting the predicting outcomes of robotic surgery videos.

In this paper, we propose a novel approach named Ternary Prior Guided Variational Autoencoder (TPG-VAE) 
for generating the future frames of dual arm robot-assisted surgical videos.
Our method combines the learned content and motion prior together with the constant class label prior
to constrain the latent space of the generation model, which is consistent with the forecasting procedure of humans by referring to various prior.
Notably, while the diversity of future tendencies is represented as distribution, the class label prior, a specific high-level target, will maintain invariant until the end of this phase, which is highly different from general robotic videos. 
Our main contributions are summarized as follows. 
\textbf{1).} A ternary prior guided variational autoencoder model is tailored for robot-assisted surgical videos. To our best knowledge, this is the first time that future scene prediction is devised for dual arm medical robots.
\textbf{2).} Given the tangled gestures of the two arms, the changeable prior from content and motion is combined with the constant prior from the class of the current action to constrain the latent space of our model. 
\textbf{3).} We have extensively evaluated our approach on the suturing task of the public JIGSAWS dataset.
Our model outperforms baseline methods in general videos in both quantitative and qualitative evaluations, especially for the long-term future.

\section{Method}


\subsection{Problem Formulation}

\begin{figure*}[t]
	\centering
	\includegraphics[width=\textwidth]{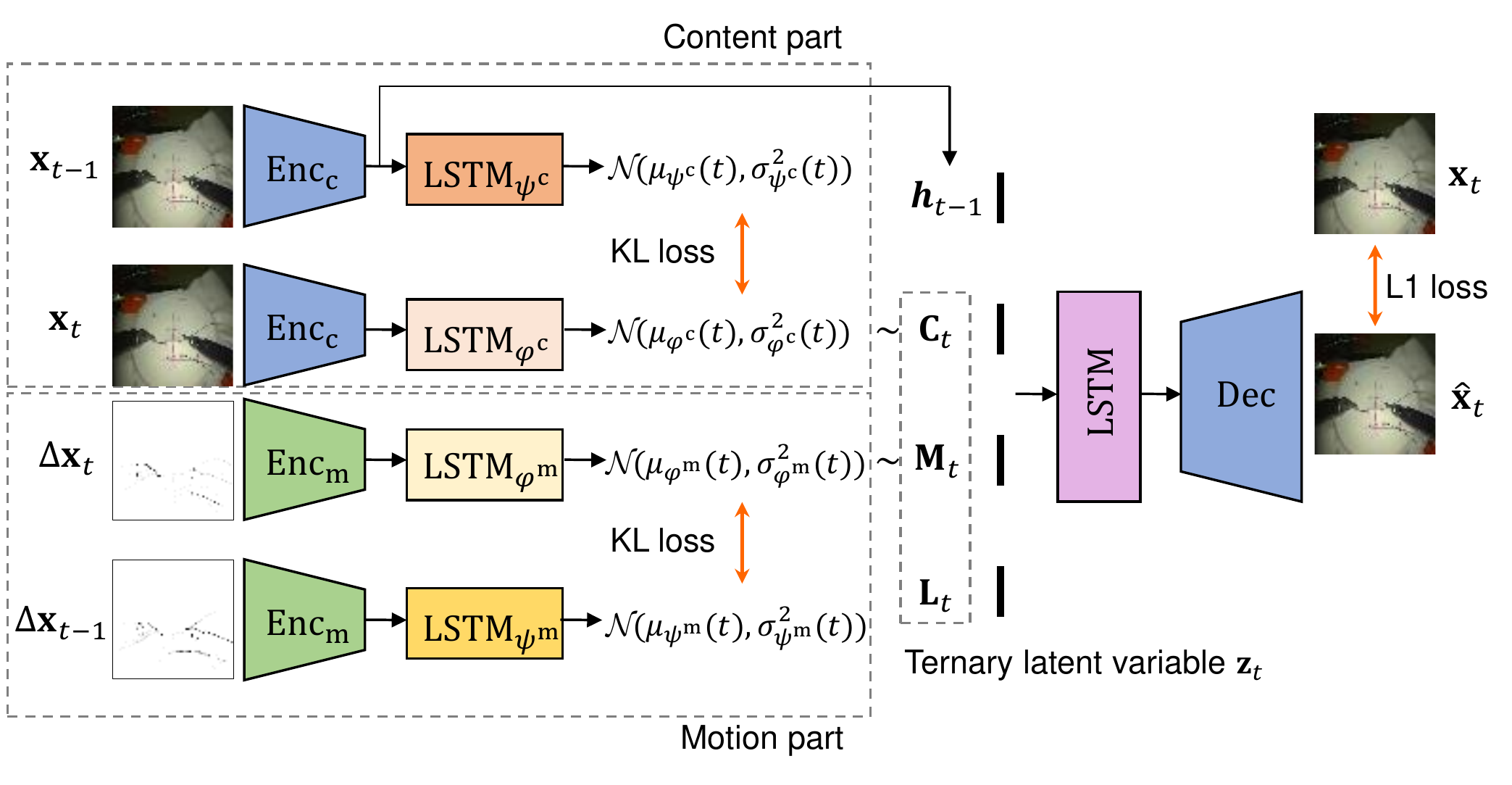}
	\caption{
		Illustration of the training process of the proposed model. 
		The posterior latent variables of content and motion at time step $t$ together with the class label prior $\textbf{L}_t$ try to reconstruct the original frame $\textbf{x}_t$ conditioning on previous frames.
		And the prior distributions from content and motion at time step $t-1$ are optimized to fit the posterior distributions at time step $t$.
	}
	\label{fig:framework}
\end{figure*}

Given a video clip $\mathbf{x}_{1:t_p}$, we aim to generate a sequence $\mathbf{\hat x}_{t_p+1:T}$ to represent the future frames, where $\mathbf{x}_t\in\mathbb{R}^{w\times h\times c}$ is the $t$-th frame with $w,~h,$ and $c$ denoting the width, height, and number of channels, respectively. 
As shown in Fig.~\ref{fig:framework}, we assume that $\textbf{x}_t$ is generated by some random process, involving its previous frame $\textbf{x}_{t-1}$ and a ternary latent variable $\textbf{z}_t$.
This process is denoted using a conditional distribution $p_\theta(\textbf{x}_t|\textbf{z}_{1:t},\textbf{x}_{1:t-1})$, which can be realized via an LSTM.
The posterior distributions of $\textbf{z}_t$ can also be encoded using LSTM.
Then, after obtaining $\textbf{z}_t$ by sampling from the posterior distributions, the modeled generation process is realized and the output $\hat{\textbf{x}}_t$ from the neural network is adjusted to fit $\textbf{x}_t$.
At the meanwhile, the posterior distribution is fitted by a prior network to learn the diversity of the future frames. 

\subsection{Decomposed Video Encoding Network}

Exacting spatial and temporal features from videos is essential for modeling the dynamics of future trends.
In this regard, we design encoders using Convolutional Neural Network (CNN) for spatial information which is followed by an LSTM to model the temporal dependency.
The content encoder $\text{Enc}_\text c$, which is realized by a CNN based on VGG net~\cite{simonyan2015very}, takes the input as the last observed frame and extracts content features as a one-dimensional hidden vector $\textbf{h}_t$.
We use $\textbf{C}_t$ to denote the latent code sampled from the content distribution, which is part of $\textbf{z}_t$. To preserve dependency, we use $\text{LSTM}_{\varphi^\text{c}}$ to model the conditional distribution of the video content. And the posterior distribution of $\textbf{C}_t$ is estimated as a Gaussian distribution $\mathcal{N}_{\varphi^\text{c}}$ with its expectation and variance as
\begin{equation}
	\begin{aligned}
	\textbf{h}_{t}=&\text{Enc}_\text c(\textbf{x}_{t}),\\
		\mu_{\varphi^\text{c}}(t), \sigma^2_{\varphi^\text{c}}(t)=&\text{LSTM}_{\varphi^\text{c}}(\textbf{h}_t). 
	\end{aligned}
\end{equation}

To further obtain an overall distribution of videos, a motion encoder is adopted to capture the changeable movements of surgical tools.
With a similar structure to the content encoder, the motion encoder $\text{Enc}_\text m$ observes the frame difference $\mathrm{\Delta}\textbf{x}_{t}$ computed by $\textbf{x}_t-\textbf{x}_{t-1}$ and outputs motion features as a one-dimensional hidden vector $\textbf{h}'_t$. 
Note that $\mathrm{\Delta}\textbf{x}_{t}$ is calculated directly using the element-wise subtraction between the two frames that are converted to gray images in advance. 
As another part of $\textbf{z}_t$, we denote the random latent code from motion as $\textbf{M}_t$.
And $\text{LSTM}_{\varphi^\text{m}}$ is utilized to calculate the posterior distribution of $\textbf{M}_{t}$ as a Gaussian distribution $\mathcal{N}_{\varphi^\text{m}}$ with its expectation and variance as
\begin{equation}
	\begin{aligned}
	\textbf{h}'_t=&\text{Enc}_\text m(\mathrm\Delta\textbf{x}_t),\\
		\mu_{\varphi^\text{m}}(t),\sigma^2_{\varphi^\text{m}}(t)=&\text{LSTM}_{\varphi^\text{m}}(\textbf{h}'_t).
	\end{aligned}
\end{equation}
It is worth to mention that our motion encoder also play a role as attention mechanism since the minor movements of instruments are caught by the difference between frames. 
The motion encoder helps alleviate the problem of limited inter-class variance which is caused by the huge proportion of unchanged parts of surgical frames.
Although our method also explicitly separates the content and motion information in a video as \cite{villegas2017decomposing}, we consider the comprehensive distribution rather than deterministic features.

For prior distributions of $\textbf{C}_t$ and $\textbf{M}_t$, two LSTMs with the same structure are applied to model them as normal distribution $\mathcal{N}_{\psi^\mathrm c}$ and $\mathcal{N}_{\psi^\mathrm m}$, respectively, as
\begin{equation}
	\begin{aligned}
		\mu_{\psi^\text{c}}(t), \sigma^2_{\psi^\text{c}}(t)&=\text{LSTM}_{\psi^\text{c}}(\textbf{h}_{t-1}),\\
		\mu_{\psi^\text{m}}(t), \sigma^2_{\psi^\text{m}}(t)&=\text{LSTM}_{\psi^\text{m}}(\textbf{h}'_{t-1}).
	\end{aligned}
	\label{eq:prior}
\end{equation}
The objective of prior networks is to estimate the posterior distribution of time step $t$ with information up to $t-1$.

\subsection{Ternary Latent Variable}

Anticipating the future development based on some inherent information can reduce the uncertainty, which also holds for the robotic video scenario.
Here, we apply the class label information of the video to be predicted as the non-learned part of the latent variable $\textbf{z}_t$, which is the available ground truth label. 
Even if there are not ground truth labels, they could be predicted since the gesture recognition problem have been solved with a relatively high accuracy~\cite{jin2017sv}, for example, 84.3\%~\cite{funke2019using} in robotic video dataset JIGSAWS~\cite{gao2014jhu,ahmidi2017dataset}.
With the obtained surgical gesture label, we encode it  as a one-hot vector $\textbf{L}_t\in\{0,1\}^{n_l}$, where $n_l$ is the number of gesture classes. 
For each video clip, $\textbf{L}_t$ is directly applied as a part of the ternary latent variable to the generation process by setting the current gesture class as 1 and all others 0.
Thus, the complete ternary latent variable of our method is written as
\begin{equation}
	\textbf{z}_t= [\textbf{C}_t,\textbf{M}_t,\textbf{L}_t],
\end{equation}
where $\textbf{C}_t$ and $\textbf{M}_t$ are sampled from $\mathcal{N}_{\varphi^{\text{c}}}$ and $\mathcal{N}_{\varphi^{\text{m}}}$ during training, respectively. Note that $\textbf{L}_t$ will keep unchanged for frames with the same class label.

After acquiring $\textbf{z}_t$, we can perform the future frame prediction using an LSTM to keep the temporal dependencies and a decoder to generate images. 
The features go through the two neural networks to produce the next frame $\hat{\textbf{x}}_t$ as
\begin{equation}
	\begin{aligned}
	\mathbf{g}_t=&\text{LSTM}_\theta (\textbf{h}_{t-1},\textbf{z}_t),\\
		\hat{\mathbf x}_t=&\text{Dec}(\textbf{g}_t).
	\end{aligned}
\end{equation}
To provide features of the static background, skip connections are employed from content encoder at the last ground truth frame to the decoder like~\cite{denton2018stochastic}. 

During inference, the posterior information of the ternary latent variable $\textbf{z}_t$ is not available and a prior estimation is needed to generate latent codes at time step $t$. 
A general way to define the prior distribution is to let the posterior distribution get close to a standard normal distribution where prior latent code is then sampled.
However, this sampling strategy tends to lose the temporal dependencies between video frames. 
Employing a recurrent structure, the conditional relationship can also be learned by a prior neural network~\cite{denton2018stochastic}. 
And we directly use the expectation of the prior distribution rather than the sampled latent code to produce the most likely prediction under the Gaussian latent distribution assumption.
Another reason is that choosing the best generation after sampling several times is not practical for online prediction scenarios.
Hence, we directly use the outputs of Eq. (\ref{eq:prior}) and $\textbf{z}_t$ is replaced of the ternary prior $\textbf{z}_t'$ as  
\begin{equation}
	\textbf{z}_t'=[\mu_{\psi^\text{c}}(t),\mu_{\psi^\text{m}}(t),\textbf{L}_t].
\end{equation}
It is worth noting that  our method considers the stochasticity during training for an overall learning, while produces the most possible generation during testing.

\subsection{Learning Process}

In order to deal with the intractable distribution of latent variables,
we train our neural networks by maximizing the following variational lower bound using the re-parametrization trick~\cite{kingma2013auto}:
\begin{equation}
	\begin{aligned}
		\sum_{t=1}^{T}&[\mathbb{E}_{q_{\varphi}(\textbf{z}_{1:T}|\textbf{x}_{1:T})}
		\log p_\theta(\textbf{x}_t|\textbf{z}_{1:t},\textbf{x}_{1:t-1})\\
		&-\beta 
		D_{\text{KL}}\left(q_{\varphi^{\mathrm c}}(\textbf{C}_t|\textbf{x}_{1: t})||p_{\psi^{\mathrm c}}(\textbf{C}_t|\textbf{x}_{1:t-1})\right)\\
		&-\beta 
		D_{\text{KL}}\left(q_{\varphi^{\mathrm m}}(\textbf{M}_t|\textbf{x}_{1: t})||p_{\psi^{\mathrm m}}(\textbf{M}_t|\textbf{x}_{1:t-1})\right)],
	\end{aligned}
\end{equation}
where $\beta$ is used to balance the frame prediction error and the prior fitting error. As shown in Fig.~\ref{fig:framework}, we use reconstruction loss to replace the likelihood term~\cite{villegas2019high} and the loss function to minimize is 
\begin{equation}
\begin{aligned}
	\mathcal{L}= \sum_{t=1}^{T}&[\|\textbf{x}_t-\hat{\textbf x}_t\|_1
	+\beta D_{\text{KL}}(\mathcal{N}_{\varphi^\text{c}}(t)||\mathcal{N}_{\psi^\text{c}}(t))
	&+\beta D_{\text{KL}}(\mathcal{N}_{\varphi^\text{m}}(t)||\mathcal{N}_{\psi^\text{m}}(t))],
\end{aligned}
\end{equation}
where $\|\cdot\|_1$ represents the $\ell_1$ loss.

\section{Experimental Results}
We evaluate our model for predicting surgical robotic motions on the dual arm \textit{da Vinci} robot system~\cite{Freschi2013}. 
We design experiments to investigate: 1) whether the complicated movements of the two robotic arms could be well predicted, 2) the effectiveness of the content and motion latent variables in our proposed video prediction model, and 3) the usefulness of the constant label prior in producing the future motions of the dual arm robots.
\subsection{Dataset and Evaluation Metrics} 
We validate our method with the suturing task of the JIGSAWS dataset~\cite{gao2014jhu,ahmidi2017dataset}, a public dataset recorded using \textit{da Vinci} surgical system. 
The gesture class labels are composed of a set of 11 sub-tasks annotated by experts.
We only choose gestures with a sufficient amount of video clips ($\sim$100), i.e. positioning needle (G2), pushing needle through tissue (G3), transferring needle from left to right (G4), and pulling suture with left hand (G6). 
The records of the first 6 users are used as training dataset (470 sequences) and the rest 2 users for testing (142 sequences), which is consistent with the leave-one-user-out (LOUO) setting in~\cite{gao2014jhu}.
Every other frame in the dataset is chosen as the input $\textbf{x}_t$.

We show quantitative comparisons by calculating VGG Cosine Similarity,  Peak Signal-to-Noise Ratio (PSNR) and structural similarity (SSIM) scores~\cite{villegas2019high} between ground truth and generated frames.
VGG Cosine Similarity uses the output vector of the last fully connected layer of a pre-trained VGG neural network. PSNR generally indicates the quality of reconstruction while SSIM is a method for measuring the perceived quality.

\subsection{Implementation Details} 
The encoder for content and decoder use the same architecture as VGG-based encoder in~\cite{denton2018stochastic}, while the motion encoder utilizes only one convolutional layer after each pooling layer with one eighth channel numbers. 
The dimensions of outputs from the two encoders are both 128.
All LSTMs have 256 cells with a single layer except $\text{LSTM}_\theta$ with two layers. 
A linear embedding layer is employed for each LSTM.
The hidden output of $\text{LSTM}_\theta$ is followed by a fully connected layer activated using a tanh function before going into the decoder, while the hidden outputs of the rest LSTMs are followed by two separate fully connected layers indicating expectation and the logarithm of variance.

Following the previous study~\cite{denton2018stochastic}, we set the resolution of the videos as $64\times64$ from $640\times480$ to save time and storage.
The dimensionalities of the $\textbf{g}_t$ and the Gaussian distributions are 128 and 16, respectively. 
We train all the components of our method using the Adam optimizer~\cite{kingma2014adam} in an end-to-end fashion, with a learning rate of $1e-4$. 
We set $\beta=1e-4$ and $T=20$, i.e., the max length of video frames to train.
For all the experiments, we train each model on predicting 10 time steps into the future conditioning on 10 observed frames, i.e. $t_p=10$. 
All models are trained with 200 epochs.

Based on their released codes, we re-implement MCnet~\cite{villegas2017decomposing} and SVG-LP~\cite{denton2018stochastic} on the robotic dataset, which are the typical methods of deterministic and stochastic predictions.
We also show the results of two ablation settings: 1) SVG-LP*: SVG-LP trained using $\ell_1$ loss and tested without sampling; 2) ML-VAE: our full model without latent variables of content.
We randomly choose 100 video clips from the testing dataset with the number of different gestures equal. 
Then, we test each model by predicting 20 subsequent frames conditioning on 10 observed frames.
The longer testing period than training demonstrates the generalization capability of each model.
For SVG-LP, we draw 10 samples from the model for each test sequence and choose the best one given each metric. 
Other VAE based methods directly use the expectation of the latent distribution without sampling for inference.

\begin{figure*}[t]
    \centering
	\includegraphics[width=\textwidth]{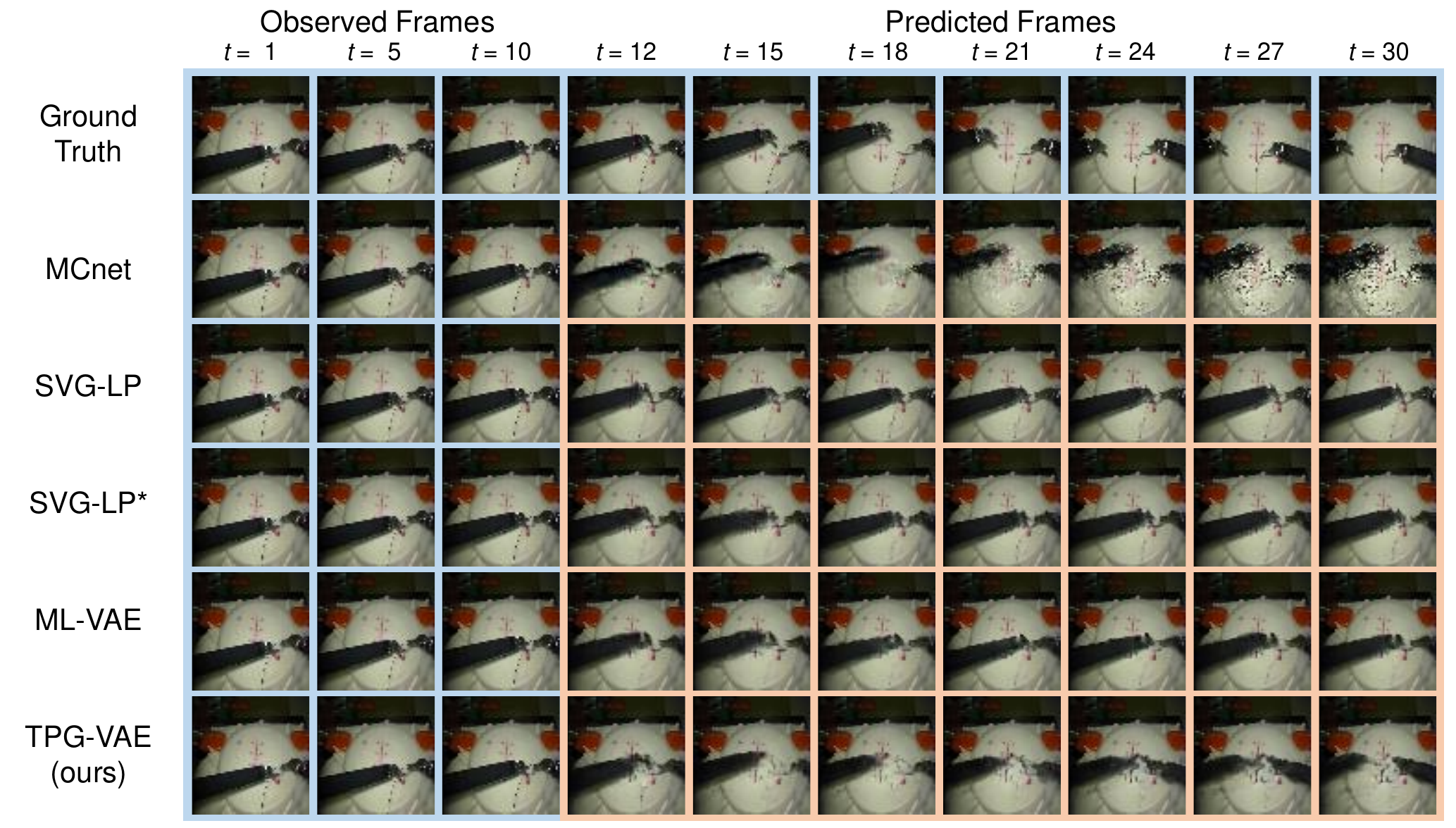}
	\caption{Qualitative results showing the gesture of G2 among different models. Compared to other VAE-based methods, our model captures the moving tendency of the left hand while other methods only copy the last ground truth frame. Frames with blue edging indicate the ground truth while the rest are generated by each model.
	}
	\label{fig:Qualitative0}
\end{figure*}

\begin{figure*}[t]
    \centering
	\includegraphics[width=\textwidth]{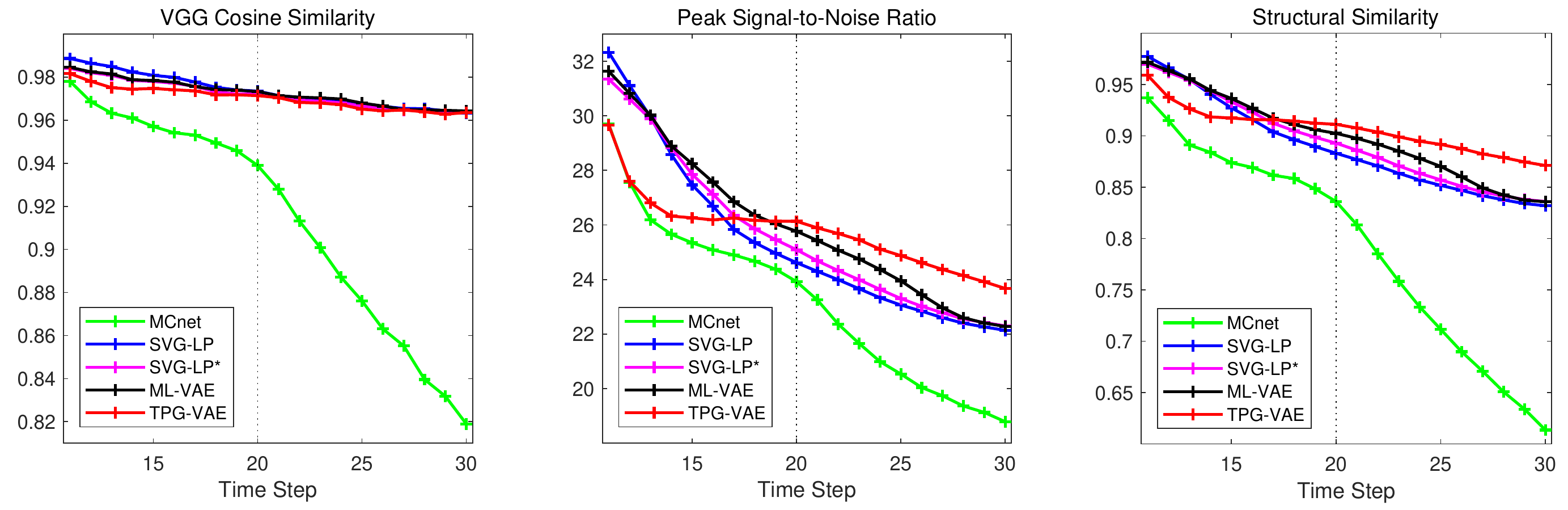}
	\caption{Quantitative evaluation on the average of the three metrics towards the 100 testing clips. 
	The dotted line indicates the frame number the models are trained to predict up to; further results beyond this line display their generalization ability.
	For the reported metrics, \textbf{higher is better}.}
	\label{fig:Quantitative}
\end{figure*}

\subsection{Results}
\subsubsection{Qualitative Evaluation}
Fig.~\ref{fig:Qualitative0} shows some of the outcomes denoting the gesture G2 from each model.
Generations from MCnet are sharp for initial time steps, but the results rapidly distort in later time steps. 
Although MCnet also utilizes skip connections, it cannot produce a crisp background in a longer time span, which implies that the back and forth moving patterns of the two arms can hardly be learned the deterministic model. 
SVG-LP and SVG-LP* tend to predict static images of the indicated gesture, which implies that the two misunderstand the purpose of the current gesture. 
ML-VAE also tends to lose the movement of the left hand, which confirms the importance of content encoder.
Capturing the movements of the two arms, our TPG-VAE model gives the closest predictions towards the ground truth. 

\subsubsection{Quantitative Evaluation}
We compute VGG cosine similarity, PSNR, and SSIM for earlier mentioned models on the 100 unseen test sequences.
Fig. \ref{fig:Quantitative} plots the average of all three metrics on the testing set.
Concerning VGG Cosine Similarity, MCnet shows the worst curve because of the lowest generation quality, while other methods behaves similarly. 
The reason might be that the frames with good quality are similar to each other on a perceptual level.  
For PSNR and SSIM, all methods maintain a relatively high level at the beginning while deteriorate as going further into the future. 
SVG-LP* shows better performance than the SVG-LP, which indicates the $\ell_1$ loss is more appropriate than MSE in this task. 
Both ML-VAE and TPG-VAE demonstrate better outcomes than the two published methods, i.e., MCnet and SVG-LP, particularly in later time steps.
Interestingly, MCnet demonstrates a poor generalization capacity that its performance curves on three metrics deteriorate faster after going through the dotted lines in Fig. \ref{fig:Quantitative}.
Our full model exhibits a stronger capability to retain image quality in longer time span.
Note that methods without sampling random variables when testing also gain favorable results, which suggests that the movements in the JIGSAWS dataset have relatively clear objectives.

\begin{table*}[t]
	\centering
	\caption{Comparison of predicted results at different time step (mean$\pm$std).}
	\label{tab:Comparison}
	\resizebox{\textwidth}{!}{
		\begin{tabular}{|l|c|c|c|c|c|c|c|c|}
			\hline
			\multirow{2}{*}{Methods} & \multicolumn{4}{c|}{PSNR} & \multicolumn{4}{c|}{SSIM} \\ \cline{2-9} 
			& t=15 & t=20 & t=25 & t=30 & t=15 & t=20 & t=25 & t=30 \\ \hline
			MCnet~\cite{villegas2017decomposing} & 25.34$\pm$2.58 & 23.92$\pm$2.46 & 20.53$\pm$2.06 & 18.79$\pm$1.83 & 0.874$\pm$0.053 & 0.836$\pm$0.058 & 0.712$\pm$0.074 & 0.614$\pm$0.073 \\
			SVG-LP~\cite{denton2018stochastic} & 27.47$\pm$3.82 & 24.62$\pm$4.21 & 23.06$\pm$4.20 & 22.14$\pm$4.09 & 0.927$\pm$0.054 & 0.883$\pm$0.080 & 0.852$\pm$0.089 & 0.832$\pm$0.088 \\ \hline
			SVG-LP* & 27.85$\pm$3.57 & 25.09$\pm$4.13 & 23.30$\pm$4.30 & 22.30$\pm$4.21 & 0.933$\pm$0.046 & 0.893$\pm$0.072 & 0.857$\pm$0.087 & 0.836$\pm$0.088 \\
			M-VAE & 27.74$\pm$3.67 & 25.14$\pm$4.09 & 23.24$\pm$4.30 & 22.15$\pm$4.10 & 0.932$\pm$0.050 & 0.894$\pm$0.072 & 0.857$\pm$0.088 & 0.834$\pm$0.087 \\
			CM-VAE & 27.44$\pm$3.83 & 25.09$\pm$4.07 & 23.02$\pm$4.19 & 22.16$\pm$4.16 & 0.927$\pm$0.056 & 0.893$\pm$0.075 & 0.853$\pm$0.087 & 0.834$\pm$0.088 \\
			CL-VAE & 28.00$\pm$3.73 & 25.32$\pm$4.15 & 23.49$\pm$4.34 & 22.24$\pm$4.28 & 0.935$\pm$0.042 & 0.897$\pm$0.073 & 0.862$\pm$0.087 & 0.835$\pm$0.088 \\
			ML-VAE & \textbf{28.24$\pm$3.51} & 25.77$\pm$4.02 & 23.95$\pm$4.26 & 22.28$\pm$4.26 & \textbf{0.936$\pm$0.046} & 0.903$\pm$0.071 & 0.870$\pm$0.084 & 0.836$\pm$0.088 \\
			\textbf{TPG-VAE (ours)} & 26.26$\pm$3.17 & \textbf{26.13$\pm$3.85} & \textbf{24.88$\pm$3.68} & \textbf{23.67$\pm$3.50} & 0.917$\pm$0.048 & \textbf{0.911$\pm$0.060} & \textbf{0.892$\pm$0.067} & \textbf{0.871$\pm$0.071} \\ \hline
		\end{tabular}
	}
\end{table*}

Table~\ref{tab:Comparison} lists the average and standard deviation of the performances of each method on the 100 testing clips.
VGG Cosine Similarity is not shown because it cannot distinguish the results of VAE based models.
All methods tend to degrade as the time step goes on.
ML-VAE also exhibits superior outcomes than the other compared methods, which verifies the effectiveness of the proposed motion encoder and class prior.
With the ternary prior as high-level guidance, our TPG-VAE maintains high generation quality while showing stable performances with the smallest standard deviation.

\subsubsection{Further Ablations}
Table~\ref{tab:Comparison} also shows additional ablations of our TPG-VAE.
Each of the following settings is applied to justify the necessity: 1) M-VAE: our full model without latent variables of content and class labels; 
2) CM-VAE: our full model without latent variables of class labels; 
3) CL-VAE: our full model without latent variables of motion.
All three ablation models give better outcomes than SVG-LP.
Comparing CM-VAE with CL-VAE, we find that class labels contribute more than motion latent variables since class information helps the model remove more uncertainty, which suggests that recognizing before predicting is a recommended choice.
To be mentioned, we do not consider the ablation setting with the only label prior since it degenerates into a deterministic model that cannot interpret the diversities of videos.

\begin{figure}[t]
	\centering
	\includegraphics[width=\textwidth]{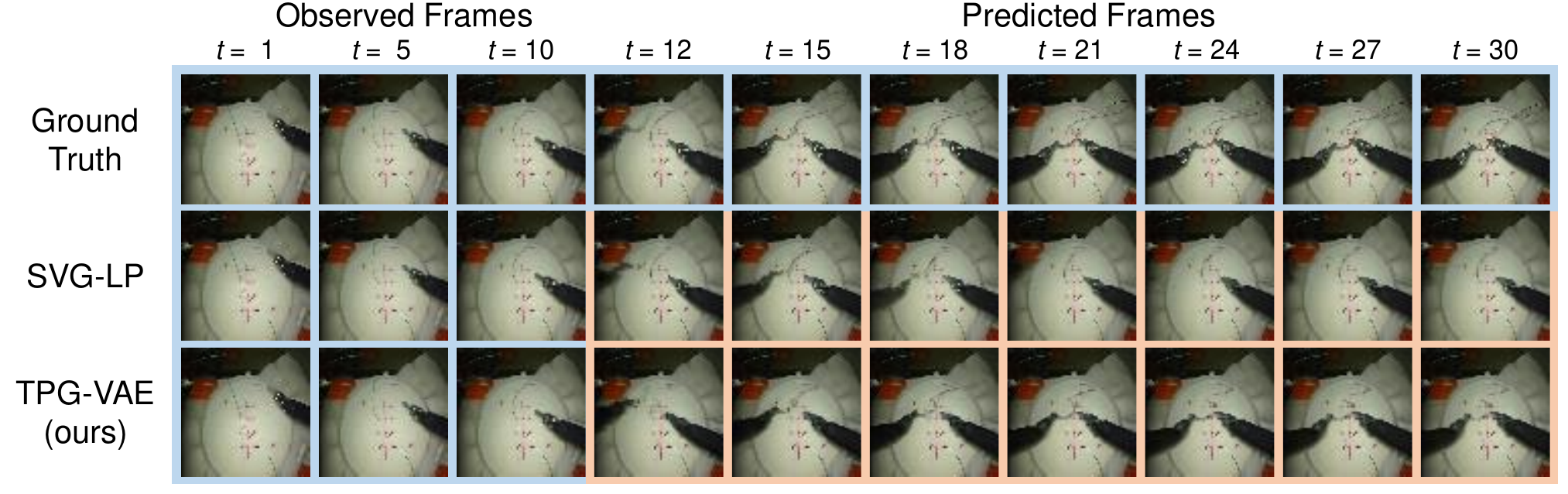}
	\caption{Qualitative comparison indicating the gesture of G4. Our method generates images with high quality and predicts the actual location of the left arm, while other methods tend to lose the left arm.
		Frames with blue edging indicate the ground truth while the rest are generated by each model.
	}
	\label{fig:Qualitative}
\end{figure}

\subsubsection{Discussion on Dual Arm Cases}
The two arms of the \textit{da Vinci} robot cooperate mutually to achieve a certain task, thus the movements are highly entangled, which makes prediction very challenging. 
Without enough prior information, the predicted frames might lead to unreasonable outcomes due to the loss of temporal consistency. 
Fig.~\ref{fig:Qualitative} shows the results of gesture G4, where the left arm is gradually getting into the visual field.
As for the first 10 predicted frames, all models realize the temporal images that the left hand is moving to the center.
For the rest 10 predictions, 
SVG-LP tend to lose the left hand due misunderstanding the current phase.
Owning to more complete guidance, i.e., the ternary prior, our TPG-VAE predicts the movements of the two arms successfully while shows crisp outcomes, which verifies our assumption that additional references help for prediction of the dual arm movements.

\section{Conclusion and Future Work}

In this work, we present a novel method based on VAE for conditional robotic video prediction, which is the first work for dual arm robots. 
The suggested model employs learned and intrinsic prior information as guidance to help generate future scenes conditioning on the observed frames.
The stochastic VAE based method is adapted as a deterministic approach by directly using the expectation of the distribution without sampling.
Our method outperforms the baseline methods on the challenging dual arm robotic surgical video dataset. 
Future work can be made to explore higher resolution generation and apply the predicted future frames to other advanced tasks.

\subsubsection{Acknowledgements.}
This work was supported by Key-Area Research and Development Program of Guangdong Province, China (2020B010165004), Hong Kong RGC TRS Project No.T42-409/18-R, National Natural Science Foundation of China with Project No. U1813204, and CUHK Shun Hing Institute of Advanced Engineering (project MMT-p5-20).



\bibliographystyle{splncs04}
\bibliography{mybibfile}

\end{document}